\documentclass{Interspeech2024}

\interspeechcameraready

\usepackage{multicol}
\usepackage{multirow}
\usepackage{amsthm}
\usepackage{graphicx}
\usepackage{subcaption}
\usepackage[table,xcdraw]{xcolor}
\usepackage{colortbl}
\usepackage{color}
\usepackage{xcolor}
\title{Acoustic Feature Mixup for Balanced Multi-aspect Pronunciation Assessment}

\name[affiliation={1}]{Heejin}{Do}
\name[affiliation={2}]{Wonjun}{Lee}
\name[affiliation={1,2}]{Gary Geunbae}{Lee}

\address{
  $^1$Graduate School of AI, POSTECH, South Korea\\
  $^2$Department of Computer Science and Engineering, POSTECH, South Korea}
\email{\{heejindo, lee1jun, gblee\}@postech.ac.kr}


\begin{document}

\maketitle

\begin{abstract}
In automated pronunciation assessment, recent emphasis progressively lies on evaluating multiple aspects to provide enriched feedback. However, acquiring multi-aspect-score labeled data for non-native language learners' speech poses challenges; moreover, it often leads to score-imbalanced distributions. In this paper, we propose two \textit{Acoustic Feature Mixup} strategies, linearly and non-linearly interpolating with the in-batch averaged feature, to address data scarcity and score-label imbalances. Primarily using goodness-of-pronunciation as an acoustic feature, we tailor mixup designs to suit pronunciation assessment. Further, we integrate fine-grained error-rate features by comparing speech recognition results with the original answer phonemes, giving direct hints for mispronunciation. Effective mixing of the acoustic features notably enhances overall scoring performances on the speechocean762 dataset, and detailed analysis highlights our potential to predict unseen distortions.

\end{abstract}

\section{Introduction}

Assisting non-native (L2) language learners to acquire foreign speaking skills, automatic pronunciation assessment is pivotal for computer-assisted pronunciation training (CAPT) systems \cite{eskenazi2009overview,franco1997automatic}. Recently, moving beyond solely evaluating phone-level scores \cite{witt2000phone,luo2009analysis,wang2012improved,shi2020context}, assessing pronunciation on multiple aspects and granularities has attracted increasing attention \cite{gong2022transformer,10095733,chao20223m, chao23_interspeech}. To achieve multi-aspect pronunciation assessment via deep learning techniques, qualified data with labeled multi-aspect scores for learner utterances is required.

However, obtaining multi-dimensional score-labeled speech data poses challenges, and score labels are prone to have imbalanced distributions \cite{do23b_interspeech, basuki2018use}, often failing to represent real-world minority cases. Such imbalanced training data skewed towards specific scores significantly degrades the model performance on samples with new or unseen score ranges \cite{do23b_interspeech}. For instance, a model trained on a biased dataset where most cases are labeled around the 2-point range may struggle to predict samples of other score ranges. Indeed, recent advancements in multi-aspect pronunciation assessment have yielded notable performance enhancements via meticulously crafted deep neural modeling \cite{gong2022transformer,10095733,chao23_interspeech} and extensive utilization of acoustic feature input \cite{chao20223m}. However, a substantial gap persists between severely score-imbalanced aspects and others, exceeding fourfold. 


In this paper, we propose two \textit{Acoustic-feature Mixup} (AM) strategies to simulate distribution shifts toward scarce positions without original speech data, thereby guiding the balanced learning for multiple scoring dimensions. Mixup \cite{zhang2017mixup} is an approach that interpolates data samples to aid in model regularization and has primarily been applied for image classification tasks \cite{uddin2020saliencymix, anonymous2023a, liu2024harnessing}. Distinct from its typical use, we suggest suitable methods for acoustic features and regression of continuous numeric labels for pronunciation assessment, where the utility is yet to be explored. In particular, we present two AM strategies: 1) static AM, which involves linear and simple combinations, and 2) dynamic AM, which integrates non-linear interpolations. Unlike existing approaches, where mixing policies are solely applied for two pairs, we consider all pairs within a batch by incorporating in-batch averaged values within the policy. 

\begin{figure}[t]
    \centering
    \includegraphics[width=7.2cm]{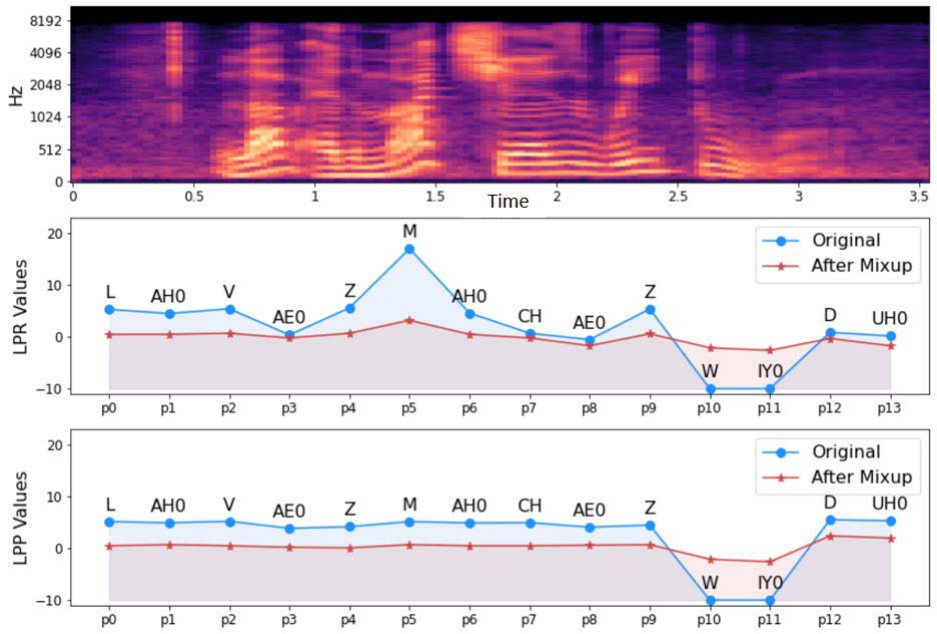}
    \caption{An example of GOP features, log phone posterior (LPP) and log posterior ratio (LPR), shift after applying dynamic Mixup.}
    \label{fig1}
\end{figure}

We mainly leverage the Goodness of Pronunciation (GOP) feature as the acoustic feature, which is determined by comparing the phone-level pronunciations of the learner and the correct answer. As GOP provides details on mispronounced phonemes, it has been widely used for pronunciation assessment. 
Our methods mix GOP features rather than the original speech data, allowing the generation of inputs that match the discriminative regions for grading without specific score-labeled speech data (Figure~\ref{fig1}). Further, we introduce multi-granular error rate features obtained from the automatic speech recognition (ASR) system. Specifically, we measure the character- and token-level match error rate between ASR results and the correct phonemes of the utterance and concatenate it with the final representation vector, thus providing direct hints for mispronunciation. Mixing up these error-rate features in parallel with GOP features further assists the model training.  

Extensive experiments on the publicly available speechocean762 dataset demonstrate the training assistance of two AM strategies on the multi-aspect pronunciation assessment framework. The original dataset exhibits severely imbalanced score distributions for aspects such as \textit{Stress} and \textit{Completeness}, a major contributor to the low performance in these aspects \cite{do23b_interspeech}. Visualizing how the proposed mixup technique shifts the existing distribution demonstrates the ability to synthesize discriminative samples. Remarkably improved performance on imbalanced aspects further suggests that AM plays a complementary role in addressing vulnerabilities in unseen score samples; thus, it assists the system in achieving aspect-wise balanced scoring.


\section{Related work}


Although multi-aspect pronunciation assessment has achieved recent success \cite{gong2022transformer,10095733,chao20223m,chao23_interspeech}, this success has been limited to aspects where the score labels of the training data are evenly distributed. The inferior performance on a specific aspect might be attributed to its highly imbalanced score-label distributions, with the majority of samples having high scores \cite{do23b_interspeech, chao23_interspeech}. 
As scores in real-world scenarios are likely to be distributed diversely, addressing such imbalances is crucial. Recent related attempts focused on training optimization by either assigning balanced weights \cite{sancinetti2022transfer} or designing balanced loss functions \cite{do23b_interspeech}. However, there has been no direct research attempting data shift, and solely optimizing training with existing data may be susceptible to potential distortion encountered in practical use. We aim to achieve robustness even with unseen range data by synthesizing data in the latent space.

Mixup \cite{zhang2017mixup} is renowned for aiding model regularization by interpolating between data samples, particularly when labeled data is scarce or not representative \cite{anonymous2023a, venkataramanan2024embedding, liu2024harnessing}. Existing studies revealed that data distribution shift effectively enhances the robustness of DNNs against adversarial samples while reducing overconfident predictions \cite{chakraborty2018adversarial, liu2021towards, venkataramanan2024embedding}. Diverse shift policies on mixups have been extensively studied for visual classification tasks \cite{kim2020puzzle, li2022openmixup, liu2022tokenmix}, but their use for pronunciation assessment has yet to be explored. Building upon these benefits, we suggest adopting a mixup for multi-aspect pronunciation assessment to overcome training difficulties induced by biased score labels.



\section{Acoustic feature mixup}
\subsection{Mixup policy}
To effectively shift the distribution of existing data skewed on specific score ranges (Figure~\ref{fig2}) and synthesize corresponding pseudo acoustic features, we introduce two AM strategies, which are static ($AM_{stat}$) and dynamic ($AM_{dyn}$). Both methods employ the average feature values of the entire samples within a mini-batch for more stabilized training; however, static AM considers simple linear transformation, while dynamic AM further incorporates non-linearity. 


\subsubsection{Static AM}
We intuitively explore a straightforward linear data transformation, which shifts the distribution in parallel. Given the $i$-th sample, where $x_i$ denotes its acoustic feature and $y_i \in \mathbb{R}^m$ represents its corresponding score vector encompassing $m$ distinct aspects, we compute the averaged acoustic feature $a_x = \frac{1}{b} \sum_{i=1}^{b} x_i$ and the averaged score label $a_y = \frac{1}{b} \sum_{i=1}^{b} y_i$ over a mini-batch of size $b$. $AM_{stat}$ linearly interpolates $x_i$ and $y_i$ with $a_x$ and $a_y$ using a mixup ratio $\lambda$ as follows:
\begin{eqnarray}
    \Tilde{x} = x - \lambda \cdot a_x \\
    \Tilde{y} = y - \lambda \cdot a_y
\end{eqnarray}
where $\lambda$ is a randomly sampled weight from a $Beta(\alpha, \alpha)$ distribution. Figure 2 illustrates that selecting lambda from a beta distribution (b) instead of a fixed constant lambda, regardless of the sample (a), helps achieve more evenly distributed pseudo labels. The synthesized pseudo acoustic feature and label pairs, $(\Tilde{x}, \Tilde{y})$, are then used for training along with the original data. Note that only mixed-up samples with labels within the range of 0 to 2 are utilized for training.
%

\subsubsection{Dynamic AM}
Emphasizing the importance of capturing intricate elements in distorted images, cutting-edge techniques for visual tasks applied dynamic mixup, which considers non-linearity existing between the samples \cite{kim2021co, anonymous2023a}. Motivated by their works and particularly tailoring for pronunciation assessment, we design a novel dynamic acoustic feature mixup policy. Specifically, we devise a non-linear interpolation between the given sample and the mini-batch mean value to shift them into a latent space. With two mixing weights $\lambda_1$ and $\lambda_2$, which are separately and randomly derived from a $Beta(\alpha, \alpha)$ distribution, the $AM_{dyn}$ is defined as follows: 
\begin{eqnarray}
    \Tilde{x} = \lambda_1 x - \lambda_2 a_x + \lambda_1 \lambda_2(x-a_x)\\
    \Tilde{y} = \lambda_1 y - \lambda_2 a_y + \lambda_1 \lambda_2(y-a_y)
\end{eqnarray}
where $x_i$, $y_i$, $a_x$, and $a_y$ are defined same as $AM_{stat}$.

\begin{figure}[t]
    \centering
    \includegraphics[width=8cm]{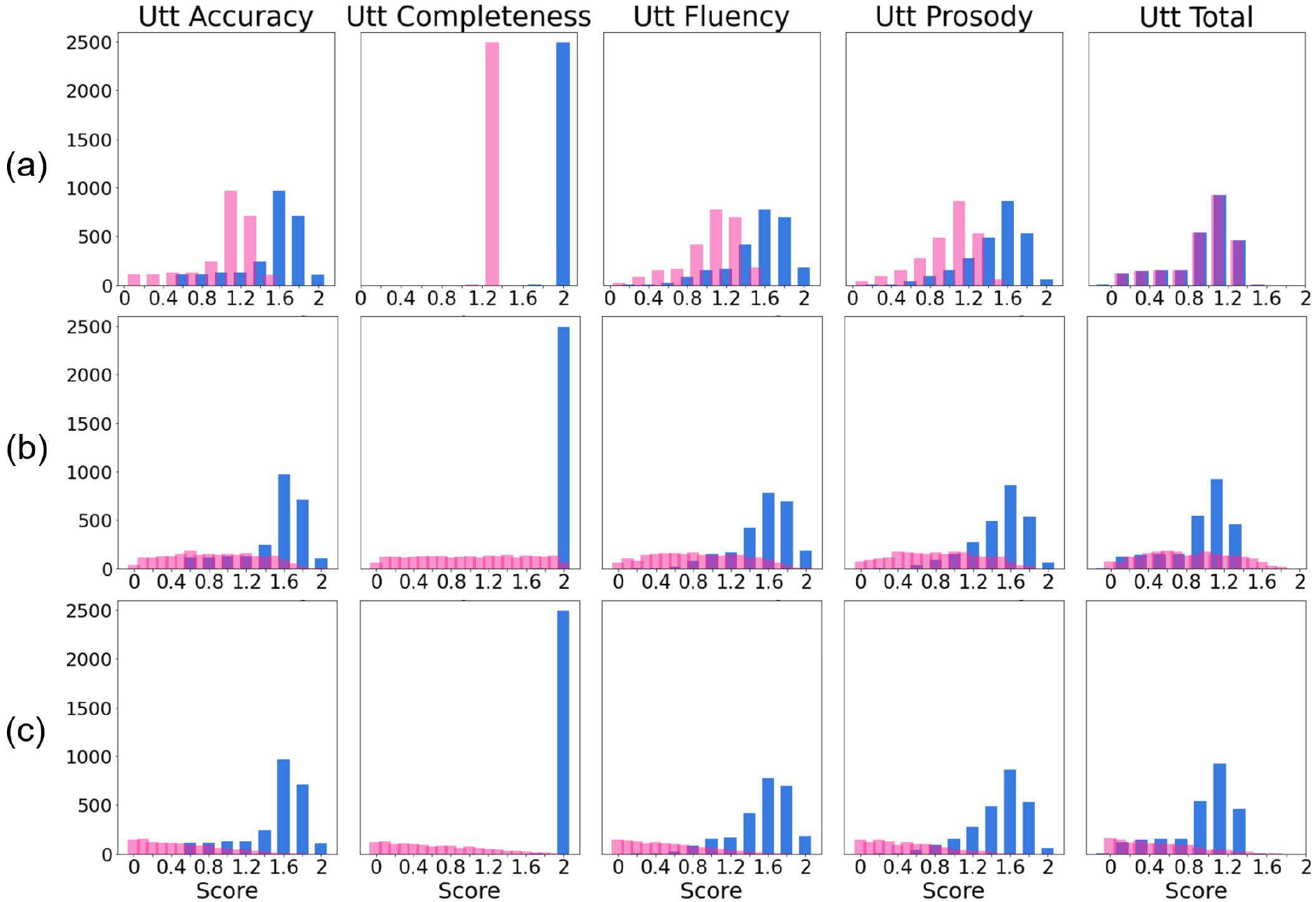}
    \caption{The utterance-level score-label distribution shift when $AM_{stat}$ with fixed $\lambda$=0.3 (a), $AM_{stat}$ with $\lambda \sim Beta(\alpha,\alpha)$ (b), and $AM_{dyn}$ (c) are applied, respectively. blue and pink bars denote original and mixed-up distribution, respectively.}
    \label{fig2}
\end{figure}

\begin{table*}[t]
\caption{\label{tab1}
Averaged MSE (for phoneme level) and PCC scores (for all levels) with standard deviation across five runs. \textbf{Acc} and \textbf{Comp} are the \textit{Accuracy} and \textit{Completeness}, respectively. GOPT{-imp} is the result of our implemented version of GOPT. \textit{+ER} denotes the addition of error-rate features. \textbf{Bold} and \underline{underline} denote the best and the second-best performance in each column, respectively.}
\centering
\scalebox{
0.83}{
\begin{tabular}{c|l|cc|ccc|ccccc}
\toprule
& & \multicolumn{2}{c|}{Phoneme Score} & \multicolumn{3}{c|}{Word Score (PCC)} & \multicolumn{5}{c}{Utterance Score (PCC)} \\
\hline
& {Model} & Acc(MSE \textbf{↓}) & Acc(PCC \textbf{↑}) & Acc \textbf{↑} & Stress \textbf{↑} & Total \textbf{↑} & Acc \textbf{↑} & Comp \textbf{↑} & Fluency \textbf{↑} & Prosody \textbf{↑} & Total \textbf{↑} \\
\hline
\multirow{6}{*}{Baseline} & \multirow{2}{*}{LSTM} & 0.089 & 0.587 & 0.511 & 0.297 & 0.524 & 0.717 & 0.123 & 0.741 & 0.744 & 0.743\\
& & \small{±0.002} & \small{±0.014} & \small{±0.014} & \small{±0.012} & \small{±0.011} & \small{±0.004} & \small{±0.143} & \small{±0.01} & \small{±0.006} & \small{±0.006} \\
& \multirow{2}{*}{GOPT} & 0.085 & 0.612 & 0.533 & 0.291 & 0.549 & 0.714 & 0.155 & 0.753 & 0.760 & 0.742\\
& & \small{±0.001} & \small{±0.003} & \small{±0.004} & \small{±0.030} & \small{±0.002} & \small{±0.004} & \small{±0.039} & \small{±0.008} & \small{±0.006} & \small{±0.005}\\
 \cline{3-12}
& \multirow{2}{*}{GOPT\small{-imp}} & 0.086 & 0.608 & 0.529 & 0.292 & 0.544 & 0.712 & 0.217 & 0.755 & 0.756 & 0.737\\
& & ±0.001 &±0.004 &±0.005 &±0.036 &±0.006 &±0.005 &±0.091 &±0.003 &±0.003 &±0.005\\
\hline\hline
\multirow{8}{*}{Ours} &  \multirow{2}{*}{\textbf{$AM_{stat}$}} & 0.085 & 0.611 & 0.532 & \textbf{0.347} & 0.551 & 0.723 & 0.281 & 0.769 & 0.766 & 0.752 \\
& & ±0.001 & ±0.007 & ±0.009 & ±0.008 & ±0.006 & ±0.007 & ±0.090 & ±0.004 & ±0.003 & ±0.007	\\
\cline{3-12}
& \multirow{2}{*}{+ER} & 0.085 & \underline{0.614} & \underline{0.538} & 0.306 & \textbf{0.558} & \underline{0.735} & \underline{0.402} & \underline{0.780} & \underline{0.779} & \underline{0.764} \\
& &  ±0.001	& ±0.005	& ±0.005	& ±0.009	& ±0.005	& ±0.001	& ±0.085	& ±0.002	& ±0.003	& ±0.005 \\
\cline{2-12}
& \multirow{2}{*}{\textbf{$AM_{dyn}$}} & 0.086 & 0.609 & 0.531 & \underline{0.332} & 0.547 & 0.726 & \textbf{0.403} & 0.769 & 0.765 & 0.753\\
& &±0.001 & ±0.007 & ±0.009 & ±0.022 & ±0.009 & ±0.003 & ±0.130 & ±0.004 & ±0.004 & ±0.003	\\
\cline{3-12}
& \multirow{2}{*}{+ER} & \textbf{0.084} & \textbf{0.617} & \textbf{0.539} & 0.317 & \underline{0.557} & \textbf{0.738} & 0.392 & \textbf{0.782} & \textbf{0.780} & \textbf{0.768}\\
& & ±0.001 & ±0.004 & ±0.003 & ±0.027 & ±0.004 & ±0.002 & ±0.182 & ±0.002 & ±0.001 & ±0.003 \\
\bottomrule
\end{tabular}}
\end{table*}

\subsection{Acoustic features}\label{sec3.2}
As the primary acoustic feature, we adopt the GOP feature instead of the original speech data. We follow the process outlined in \cite{hu2015improved, gong2022transformer} for GOP feature generation. Specifically, the speech audio and its canonical transcription are first given to the acoustic model, yielding a sequence of phonetic posterior probabilities. Subsequently, following phoneme-level force alignment, these probabilities are converted into 84-dimensional GOP features. The dimensionality 84 stems from the concatenation of log phone posterior (LPP) and log posterior ratio (LPR), each comprising 42 dimensions, calculated for each of the 42 source phones within the Librispeech acoustic model.
The LPP of a phone $\varphi$ and LPR of observing phone $\varphi_j$ given phone $\varphi_i$ are defined as follows \cite{gong2022transformer}:
\begin{align} 
LPP(\varphi) & \approx\frac{1}{t_e-t_s+1}\sum_{t_s}^{t_e} \mathrm{log}\ p(\varphi|o_t) \\
LPR(\varphi_j|\varphi_i) & =\mathrm{log}\ p(\varphi_j|o;t_s,t_e) - \mathrm{log} p(\varphi_i|o;t_s,t_e) 
\end{align} 
where $o_t$ is the input observation of the frame $t$, and the start and end frame indexes are $t_s$ and $t_e$, respectively.

In addition, we incorporate fine-grained error rate features to provide the model with direct information about mispronunciations. Considering that correct phonemes for the utterances learners need to mimic are provided, we compare the learner's ASR-hypothesized phonemes to the reference answer phonemes to extract the error rate. Specifically, we use the character error rate (CER) and the match error rate (MER). CER is measured by dividing the number of missed characters by the number of characters in the reference. MER is calculated by dividing the number of missed tokens (phonemes in our work) by the total number of tokens in the union of the hypothesis and reference. While CER focuses on individual character errors, MER focuses on correct phoneme matches. The extracted error rates are concatenated with the model representation before passing to the final linear layer for each aspect score prediction.



\subsection{Loss function}
For training, we employ the mean squared error (MSE) loss, a widely utilized function for the pronunciation assessment task \cite{gong2022transformer,10095733, chao20223m}. The overall loss is determined by aggregating the individual losses at each granularity level, where each loss represents the multi-aspect-averaged value within that level. The total loss is defined as follows:
\begin{small}\begin{equation}
\label{eq20}
    MSE_{total} = \sum^{M}\frac{1}{N}\sum_{}^{N}MSE_{mn}
\end{equation}\end{small}
given the $M$ granularity levels and $N$ aspects. In this work, 3 levels of granularity and 9 aspects are applied.

\begin{table*}[t]
\caption{\label{tab2}
Comparison of results between using a fixed lambda value of 0.3 in $AM_{stat}$ (fix) and using random weights following a beta distribution (beta). \textit{+ER} denotes the addition of error-rate features.}
\centering
\scalebox{
0.83}{
\begin{tabular}{l|cc|ccc|ccccc}
\toprule
& \multicolumn{2}{c|}{Phoneme Score} & \multicolumn{3}{c|}{Word Score (PCC)} & \multicolumn{5}{c}{Utterance Score (PCC)} \\
\hline
{Model} & Acc(MSE ↓) & Acc(PCC ↑) & Acc ↑ & Stress ↑ & Total ↑ & Acc ↑ & Comp ↑ & Fluency ↑ & Prosody ↑ & Total ↑ \\
\hline
 \multirow{2}{*}{\textbf{$AM_{stat}$}(fix) +ER} & 0.085 & 0.614 & 0.537 & \textbf{0.324} & 0.555 & {0.736} & 0.302 & 0.780 & {0.780} & {0.766} \\
& ±0.001 & ±0.004 & ±0.004 & ±0.025 & ±0.003 & ±0.007 & ±0.054 & ±0.004 & ±0.003 & ±0.007 \\
\hline
 \multirow{2}{*}{\textbf{$AM_{stat}$}(beta) +ER} & 0.085 & {0.614} & {0.538} & 0.306 & {0.558} & 0.735 & \textbf{0.402} & {0.780} & {0.779} & 0.764 \\
&  ±0.001	& ±0.005	& ±0.005	& ±0.009	& ±0.005	& ±0.001	& ±0.085	& ±0.002	& ±0.003	& ±0.005 \\
\bottomrule
\end{tabular}}
\end{table*}

\section{Experiments}
We evaluate our $AM$ methods on the open-source speechocean762 (\cite{zhang2021speechocean762}) dataset, which includes the speech data of non-native language learners and the corresponding labeled multi-aspect scores. While its multifaceted labeled scores on multi-granular levels provide diverse opportunities for the multi-aspect pronunciation assessment, they have severely imbalanced labels, particularly for specific aspects. The dataset comprises 2500 utterances of training and test sets, respectively. We employ the fundamental framework, the GOPT \cite{gong2022transformer} model, for training to explore the sole effects of the mixup itself without supplementary modeling techniques. GOPT is based on a Transformer \cite{vaswani2017attention} encoder and utilizes the 84-dimensional GOP features obtained with the process described in Section~\ref{sec3.2}. The GOP features are first projected to 24 dimensions by a projection layer and combined with canonical phoneme and positional embedding. Then, the combined input is fed into a three-layer transformer encoder with 24 embedding dimensions. 

To ensure a fair comparison, we kept all settings except those related to the proposed method and GPU identical to the GOPT. Specifically, using the Adam optimizer, we set the learning rate as 1e-3 and batch size as 25 on 100 epoch training. For the acoustic model\footnote{\url{https://kaldi-asr.org/models/m13}} to obtain GOP features, we used the LibriSpeech \cite{panayotov2015librispeech} 960-hour data-trained model. $\alpha$ for beta distribution is set as 1 to create even likelihoods for mixing coefficients. To acquire error-rate features, we employed a wav2vec2.0 with 315 million parameters \cite{baevski2020wav2vec} as the ASR model. For phoneme transcription and evaluation, we aligned the ASR model's vocabulary with the speechocean762 dataset and trained the ASR model with the CTC head \cite{graves2012connectionist}. 
GTX 2080Ti GPU is used, and the averaged PCC results of five distinct runs are reported with the standard deviation. Following prior studies, MSE is also used to measure phoneme-level accuracy.

\section{Results and discussion}
\subsection{Main result}
The main results presented in Table~\ref{tab1} highlight the effectiveness of both our $AM_{stat}$ and $AM_{dyn}$ methods in improving the training of the DNN-based model across multiple aspects at the phoneme, word, and utterance levels. Particularly noteworthy is the approximately 25\% enhancement in assessment performance for the previously weakest aspect, \textit{Completeness}, indicating a more balanced outcome across various aspects. Also, improvements are observed for the \textit{Stress}, another highly imbalanced aspect, but the extents are not as significant. Notably, \textit{Completeness} is scored on a continuous scale from 0 to 10, while \textit{Stress} is scored on a scale of either 5 or 10. Therefore, our method of smoothly shifting the distribution to achieve evenness might be more suitable for the former. 

Overall, $AM_{dyn}$+ER exhibits the highest performance tendency, followed closely by $AM_{stat}$+ER. While pseudo labels generated by static mixup span the entire score spectrum (Figure~\ref{fig2}; b), those from dynamic mixup tend to be distributed more on rare or lower scores (Figure~\ref{fig2}; c); thus, higher results on $AM_{dyn}$+ER imply its potential guidance for more adversarial synthesis. A noticeable point is made for severely imbalanced and inferior aspects such as \textit{Completeness} and \textit{Stress}: excluding error-rate features in static and dynamic mixups yields better performance. This discrepancy could be attributed to ER's reliance on ASR model results, which may propagate ASR errors during the mixing process, unlike fixed and reliable human-annotated score labels. 

\begin{table}[t]
\caption{\label{tab3} Ablation results in error-rate features. The multi-aspect averaged performances within each level are reported.}
\centering
\scalebox{
0.83}{
\begin{tabular}{l|cc|c|c}
\toprule
&  \multicolumn{2}{c|}{Phoneme} & Word & Utterance \\
\hline
Model & MSE ↓ & PCC ↑ & Avg PCC↑ & Avg PCC↑\\
\hline
GOPT & 0.085 & 0.612 & 0.458 & 0.625 \\
\hline
CER & 0.086 & 0.610 & 0.453 & 0.637 \\
MER & 0.085	& 0.613 & 0.456 & 0.650 \\
CER + MER & 0.086 & 0.612 & 0.462 & 0.656 \\
\hline
\rowcolor{lightgray!20}
\textbf{+$AM_{stat}$} & \textbf{0.085} & \textbf{0.614} & \textbf{0.467} & \textbf{0.692} \\
\rowcolor{lightgray!20}
\textbf{+$AM_{dyn}$} & \textbf{0.084} & \textbf{0.617} & \textbf{0.471} & \textbf{0.692} \\
\bottomrule
\end{tabular}}
\end{table}

\subsection{Mixup weight choices}
We investigate the impact of the choice of mixture ratio in static AM, whether to set it to a fixed value or follow a random beta distribution. When weights are fixed at a static value of 0.3, the shifted distribution of labels appears quite rigid  (Figure~\ref{fig2}; a). However, the superior performance of the fixed $AM_{stat}$ in word-level \textit{Stress} as shown in Table~\ref{tab2} suggests that such rigidity might be advantageous in discrete aspects. Conversely, the contrasting trend observed in \textit{Completeness} indicates that a smoother shift could be beneficial for aspects requiring continuous predictions.


\subsection{Error rate ablation studies}
We conduct extensive ablation studies to examine the individual and combined effects of each error rate on model training. The results in Table 3 indicate that, when used individually, MER has a greater impact than CER. Particularly at the utterance level, MER proves beneficial, likely due to its measurement method focusing on phonemes across the entire utterance. Notably, while neither individually aids at the word level, their combined usage shows performance improvement, indicating a synergistic effect between the two error factors. Moreover, the inclusion of +$AM_{stat}$ and +$AM_{dyn}$, which incorporate original and mixed-up error rates into the final model vector, remarkably improves the PCC across all levels, highlighting the effectiveness of auxiliary combining ER features.


\subsection{Mixup direction matters}
\begin{figure}[t]
    \centering
    \includegraphics[width=7.8cm]{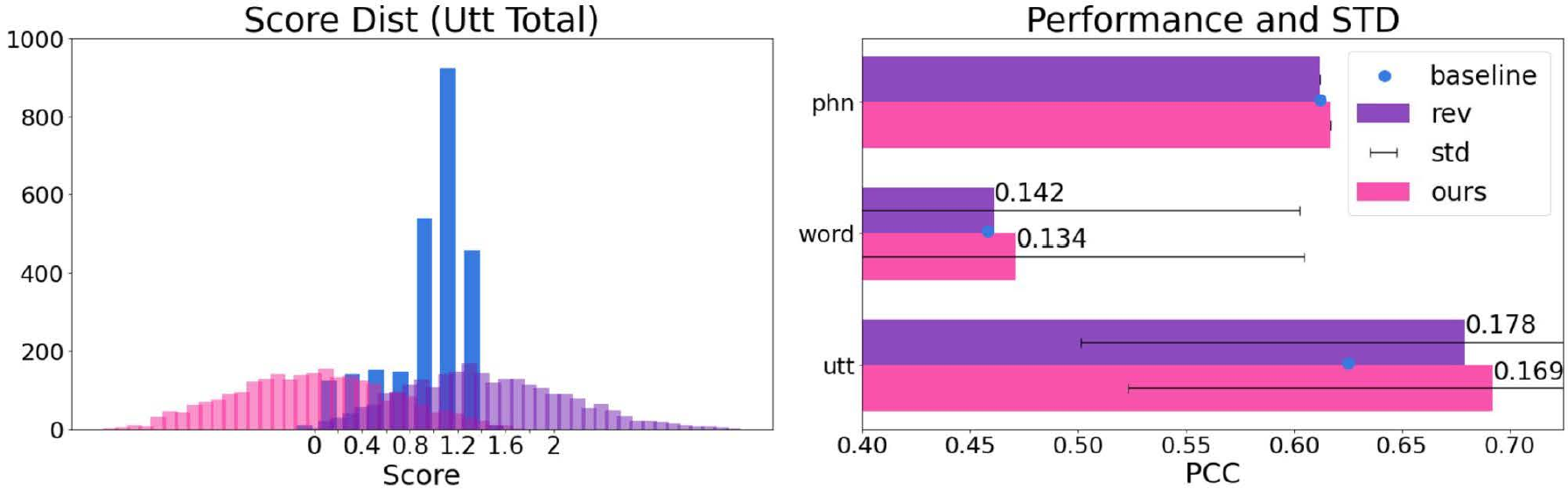}
    \caption{Score-label distribution shift when $AM_{dyn}$ is applied with the {\color{magenta}\textbf{original}} and the {\color{violet}\textbf{reverse}} directions (left), and PCC performance and standard deviation of PCC of aspects within each granularity level (right).}
    \label{fig3}
\end{figure}

We further analyze whether our hypothesized shift toward underserved areas is indeed beneficial compared to the opposite direction. In particular, we adjust our formula from the original ($\Tilde{x} = \lambda_1 x - \lambda_2 a_x + \lambda_1 \lambda_2(x-a_x)$) to the following ($\Tilde{x} = \lambda_1 x + \lambda_2 a_x + \lambda_1 \lambda_2(x-a_x)$), aiming to move in a direction proportional to the average score, inspired by \cite{anonymous2023a}. We call this as \textit{reversed} $AM_{dyn}$. In the left part of Figure 3, we observe that the \textit{reversed} $AM_{dyn}$ indeed induces shifts in the opposite direction as intended. This suggests that while the original $AM_{dyn}$ generates minority samples more frequently, the reversed $AM_{dyn}$ favorably synthesizes majority samples. An interesting finding is that $AM_{dyn}$ outperforms reversed $AM_{dyn}$ across all granularity levels (Figure~\ref{fig3}; bar charts), with even the decreasing PCC standard deviation among the aspects within each level. The result reveals that our approach not only contributes to achieving competitive performance but also facilitates balanced learning across overall aspects as we intended.


\section{Conclusion}
In this work, we propose two \textit{Acoustic Feature Mixup} strategies, $AM_{stat}$ and $AM_{dyn}$, which consider linear and non-linear interpolation between the samples and in-batch averaged feature, respectively. Primarily leveraging the GOP features but additionally introducing the error rate features, we design effective mixup policies. To evaluate our method on the DNN-based model, we use the foundational system for the multi-aspect pronunciation assessment task. Experiments with the highly imbalanced speechocean762 dataset exhibit overall performance improvement across all aspects, demonstrating our assistance in balanced scoring. Extensive analysis further demonstrates the potential for our smoother shift with $AM$ to enhance prediction for adversarial or unseen samples.

\section{Acknowledgements}
This research was partly supported by the MSIT (Ministry of Science and ICT), Korea, under the ITRC (Information Technology Research Center) support program (IITP-2024-2020-0-01789) supervised by the IITP (Institute for Information \& Communications Technology Planning \& Evaluation) and by Institute of Information \& communications Technology Planning \& Evaluation (IITP) grant funded by the Korea government (MSIT) (No.2022-0-00223, Development of digital therapeutics to improve communication ability of autism spectrum disorder patients).
\bibliographystyle{IEEEtran}
\bibliography{main}

\end{document}